%% file: main.tex
\input{misc/packages}
\begin{document}

\title{Spiking Neural Networks for Low-Power Vibration-Based Predictive Maintenance
}

\author{
    \IEEEauthorblockN{
        Alexandru Vasilache$^{1,2}$, Sven Nitzsche$^{1,2}$, 
        Christian Kneidl$^{3}$, Mikael Tekneyan$^{3}$, 
        Moritz Neher$^{1,2,4}$, \\
        Juergen Becker$^{2}$
    }
    
    \IEEEauthorblockA{
        $^{1}$ \textit{FZI Research Center for Information Technology, Karlsruhe, Germany}\\ 
        $^{2}$ \textit{Karlsruhe Institute of Technology, Karlsruhe, Germany} \\
        $^{3}$ \textit{NETZSCH Pumpen und Systeme GmbH, Waldkraiburg, Germany} \\
        $^{4}$ \textit{Infineon Technologies, Dresden, Germany} \\
        \email{vasilache@fzi.de}\orcidlink{0009-0007-7722-5127}   
    }
    
}

\maketitle
\input{misc/acronyms}
\input{sections/abstract}


\section{Introduction}
    \input{sections/introduction}

\section{Use Case}
    \input{sections/usecase}

\section{Methods}
    \input{sections/methods}


\section{Results}
    \input{sections/results}


\section{Conclusion and Outlook}
    \input{sections/conclusion}

\section*{ACKNOWLEDGMENTS} This research is funded by the German Federal Ministry of Education and Research as part of the project ”ThinKIsense“, funding no. 16ME0564.

\bibliographystyle{IEEEtran}
\bibliography{refs}

\end{document}

%% file: misc/packages.tex
\documentclass[conference]{IEEEtran}
\usepackage{cite}
\usepackage{amsmath,amssymb,amsfonts}
\usepackage{algorithmic}
\usepackage{graphicx}
\usepackage{textcomp}
\usepackage{xcolor}
\usepackage{hyperref}
\usepackage{caption}
\usepackage{subcaption}
\usepackage[per-mode=fraction]{siunitx}
\graphicspath{ {./figs/} }
\def\BibTeX{{\rm B\kern-.05em{\sc i\kern-.025em b}\kern-.08em
    T\kern-.1667em\lower.7ex\hbox{E}\kern-.125emX}}

\usepackage[nolist]{acronym}
\usepackage{orcidlink}
\usepackage[english]{babel}
\usepackage{randtext}
\newcommand{\email}[1]{
  \texttt{\randomize{#1}}
}

\usepackage{booktabs} 
\usepackage{multirow} 

%% file: misc/acronyms.tex
\begin{acronym}[Longest Abrev] 
\acro{ANN}{Artificial Neuronal Network}
\acro{AE}{Auto-Encoder}
\acro{CNN}{Convolutional Neural Network}
\acro{DNN}{Deep Neural Network}
\acro{LIF}[LIF]{Leaky Integrate-and-Fire}
\acro{IF}[IF]{Integrate-and-Fire}
\acro{LI}[LI]{Leaky Integrate}
\acro{ALIF}{Adaptive Leaky Integrate-and-Fire}
\acro{adex}[AdEx]{adaptive exponential integrate-and-fire}
\acro{BLE}{Bluetooth Low Energy}
\acro{LSTM}{Long Short-Term Memory}
\acro{RNN}{Reccurent Neural Network}
\acro{LSNN}{Long Short-Term Spiking Neural Network}
\acro{NN}{Neural Network}
\acro{SNN}{Spiking Neural Network}
\acro{ML}{Machine Learning}
\acro{DL}{Deep Learning}
\acro{AI}{artificial intelligence}
\acro{DFT}{Discrete Fourier Transform}
\acro{FFT}{Fast Fourier Transform}
\acro{STFT}{Short-Time Fourier transform}
\acro{BPTT}{Backpropagation Through Time}
\acro{BP}{Backpropagation}
\acro{LMD}{Local Mean Decomposition}
\acro{GRF}{Gaussian Receptive Field}
\acro{TTFS}{Time-to-First-Spike}
\acro{PCA}{Principal component analysis}
\acro{FPGA}[FPGA]{Field Programmable Gate Array}
\acro{SF}[SF]{Step-Forward}
\acro{SW}[SW]{Sliding Window}
\acro{MW}[MW]{Moving Window}
\acro{BSA}[BSA]{Ben's Spiker Algorithm}
\acro{PWM}[PWM]{Pulse Width Modulation}
\acro{PWMB}[PWMB]{Pulse Width Modulated-Based}
\acro{RL}[RL]{Reinforcement Learning}
\acro{FIR}[FIR]{Finite Impulse Response}
\acro{MSE}[MSE]{Mean Squared Error}
\acro{GRF}[GRF]{Gaussian Receptive Field}
\acro{CE}[CE]{Cross-Entropy}
\acro{MAE}{Mean Absolute Error}
\acro{MAPE}{Mean Absolute Percentage Error}
\acro{MRPE}{Mean Relative Percentage Error}
\acro{FPR}{False Positive Rate}
\acro{FNR}{False Negative Rate}
\acro{AUC}{Area Under the Curve}
\acro{FSE}{Full Scale Error}
\acro{MVE}{Measured Value Error}
\acro{PM}{Predictive Maintenance}
\acro{IIoT}{Industrial internet of things}
\acro{PCP}{Progressing Cavity Pump}
\acro{MAC}{Multiply-Accumulate Operation}
\end{acronym}

%% file: sections/abstract.tex
\begin{abstract}
Advancements in Industrial Internet of Things (IIoT) sensors enable sophisticated Predictive Maintenance (PM) with high temporal resolution. For cost-efficient solutions, vibration-based condition monitoring is especially of interest. However, analyzing high-resolution vibration data via traditional cloud approaches incurs significant energy and communication costs, hindering battery-powered edge deployments. This necessitates shifting intelligence to the sensor edge. Due to their event-driven nature, Spiking Neural Networks (SNNs) offer a promising pathway toward energy-efficient on-device processing. 
This paper investigates a recurrent SNN for simultaneous regression (flow, pressure, pump speed) and multi-label classification (normal, overpressure, cavitation) for an industrial progressing cavity pump (PCP) using 3-axis vibration data.
Furthermore, we provide energy consumption estimates comparing the SNN approach on conventional (x86, ARM) and neuromorphic (Loihi) hardware platforms. Results demonstrate high classification accuracy (\textgreater97\%) with zero False Negative Rates for critical Overpressure and Cavitation faults. Smoothed regression outputs achieve Mean Relative Percentage Errors below 1\% for flow and pump speed, approaching industrial sensor standards, although pressure prediction requires further refinement. Energy estimates indicate significant power savings, with the Loihi consumption (0.0032 J/inf) being up to 3 orders of magnitude less compared to the estimated x86 CPU (11.3 J/inf) and ARM CPU (1.18 J/inf) execution. Our findings underscore the potential of SNNs for multi-task PM directly on resource-constrained edge devices, enabling scalable and energy-efficient industrial monitoring solutions.
\end{abstract}

%% file: sections/introduction.tex
\label{sec:introduction}

The integration of sensors within the \ac{IIoT} enables advanced \ac{PM} strategies, crucial for optimizing industrial processes and preventing costly downtimes \cite{mckinsey2021_pm_study}. Condition monitoring often relies on vibration data, particularly for rotating machinery like pumps. However, acquiring high-resolution vibration data generates substantial data volumes, and traditional approaches involving data transmission to the cloud for analysis incur significant energy costs, limiting the operational lifetime of battery-powered sensors \cite{nitzsche_ultra-low_2021}. Addressing these limitations necessitates shifting \ac{AI}-driven analysis directly to the sensor edge.

\acp{SNN} offer a compelling solution for low-power edge \ac{AI}. Their event-driven computation holds the potential for significant energy savings compared to traditional \acp{ANN}, especially on specialized neuromorphic hardware \cite{blouw2019_snn_energy}. Research applying \acp{SNN} to \ac{PM} tasks, surveyed in \cite{vasilache2023survey}, often focuses on fault detection in components like bearings \cite{nitzsche_ultra-low_2021, dennler_online_2021, zuo_spiking_2021, li_research_2022, zuo_multi-layer_2022} or gears \cite{ali_novel_2022}. Common approaches involve converting vibration data to spikes using time-frequency transformations and various encoding schemes (e.g., Population Coding, TTFS, Current Injection). While shallow feed-forward \ac{LIF} networks are prevalent, more complex architectures like \acp{LSNN} \cite{zanatta_damage_2021, barchi_spiking_2021} or Reservoir SNNs \cite{dey_efficient_2022, kholkin_comparing_2023} are emerging. Training is mostly supervised, using surrogate gradients or ANN-to-SNN conversion. However, a notable gap exists, as most studies do not report concrete hardware implementations or  energy estimates on edge platforms \cite{vasilache2023survey, dennler_online_2021, barchi_spiking_2021}.

Among other goals, this paper aims to bridge that gap by presenting energy estimates of a recurrent \ac{SNN}, developed for a complex, multi-task \ac{PM} problem on an industrially relevant use case. Using 3-axis vibration data, we target simultaneous regression (flow, pressure, pump speed) and multi-label condition classification (normal, overpressure, cavitation). Section~\ref{sec:use_case} presents the targeted use case and data acquisition process. In Section~\ref{sec:methods}, we present the complete methodology, encompassing data preprocessing, spike encoding, the recurrent \ac{SNN} architecture, training procedures, hyperparameter optimization, compression techniques (pruning, quantization), and evaluation approach. Finally, Section~\ref{sec:results} shows performance evaluation and energy estimations comparing conventional (x86 CPU, ARM CPU) and neuromorphic (Loihi) platforms.

%% file: sections/usecase.tex
\label{sec:use_case}

Due to the lack of a standard benchmark dataset in the context of \ac{PM} or \acp{PCP}, a custom dataset was recorded specifically for this purpose. What follows is the description of the data acquisition process.

\subsection{Pump System} 

\begin{figure}[t]
  \centering
  \includegraphics[width=0.48\textwidth]{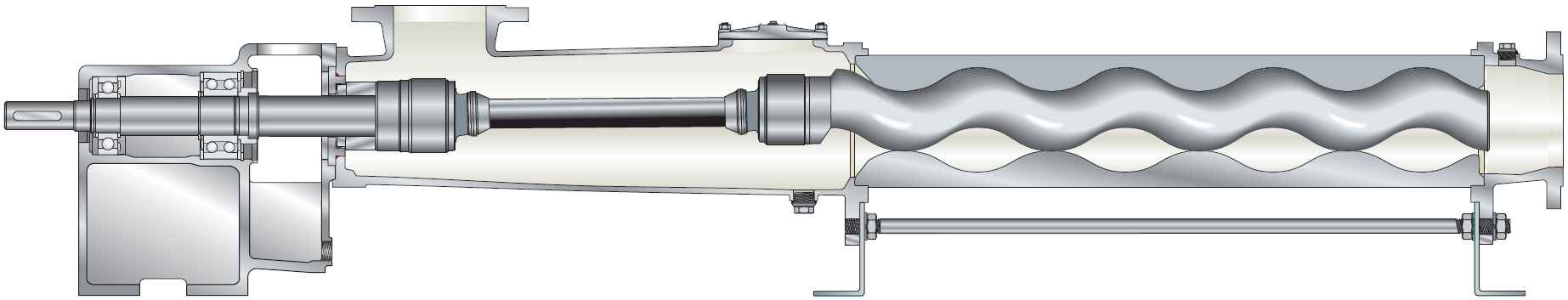}
  \caption{Progressing cavity pump (PCP)}
  \label{fig:netzsch_pumpe_sy}
\end{figure}

Throughout this study, the NETZSCH NM021BY02S \ac{PCP} \cite{netzsch_pump} was used, which is a type of rotary positive displacement pump \cite{Moineau1932}. The pump's primary components include an elastomer stator, a metal screw rotor, and a joint bar (see Figure~\ref{fig:netzsch_pumpe_sy}).
Its distinctive feature is the eccentric and rotating motion of the rotor within the stator, which results in pumping chambers sealed through pretension.
Predominantly utilized for handling viscous media and in dosing applications, the \ac{PCP} is versatile across a broad spectrum of uses.

To avoid damage in \acp{PCP}, it is necessary to predict deviations from normal operation conditions, considered impermissible, which either cause immediate damage or lead to long-term degradation of pump components. These deviations can be monitored through objective parameters, such as pump speed (rotational speed of the rotor), inlet pressure, outlet pressure, and flow rate. They include, among others, overpressure and cavitation.

The investigated NETZSCH pump is designed around a maximum discharge pressure of 12 bar. Any recordings exceeding this limit, including a small transition zone, were labeled as overpressure.
    
Any conditions with a suction pressure below 0.75 bar were labeled as cavitation, as the pump setup was known to consistently cavitate below this operation point. In parallel, an experienced operator adjusted the labels by listening with a stethoscope for the specific sound of cavitation.

As such, for each 3-axis vibration recording of 16384 points measured at 6664 Hz, a label was assigned based on the measured regression (flow, outlet pressure, pump speed) and classification (normal, overpressure, cavitation) values.

\subsection{Sensors} 

To measure the vibration of the pump, the SIEMENS SITRANS MS200 \cite{siemens_sistrans} was employed. The vibration sensor was mounted at the hub position on the pump, mechanically attached using M8 threaded stud bolts. It features an integrated \ac{BLE} interface, allowing transmission of the collected sensor data to a central gateway. With a sampling rate of 6664 samples per second, the sensor offers high temporal resolution, making it ideal for detecting subtle changes in the vibration behavior of machinery.

For pressure measurement, the IFM PG2453 sensor \cite{imf_pg2453_nodate} was used, operating in the -1 -- 25 bar range and having an error within +/- 0.2\% of the measurement range.

The Endress \& Hauser Promag W 5W3B25-5TT1/0 \cite{endressandhauser_promag_nodate} was used to measure the flow, detecting values between 0 and 18 $m^3/h$ with an error of +/- 0.5\% from the measured value.

To measure pump speed, the Omron E2B-M12KN05-M1-B1 \cite{omron_e2b-m12kn05-m1-b1_nodate} was used. Its range is between 0 and 2000 $\text{min}^{-1}$, and offers an accuracy of +/- 0.1\% from the measured value.

\subsection{Data Generation} 

\begin{figure}[t]
  \centering
  \includegraphics[width=0.48\textwidth]{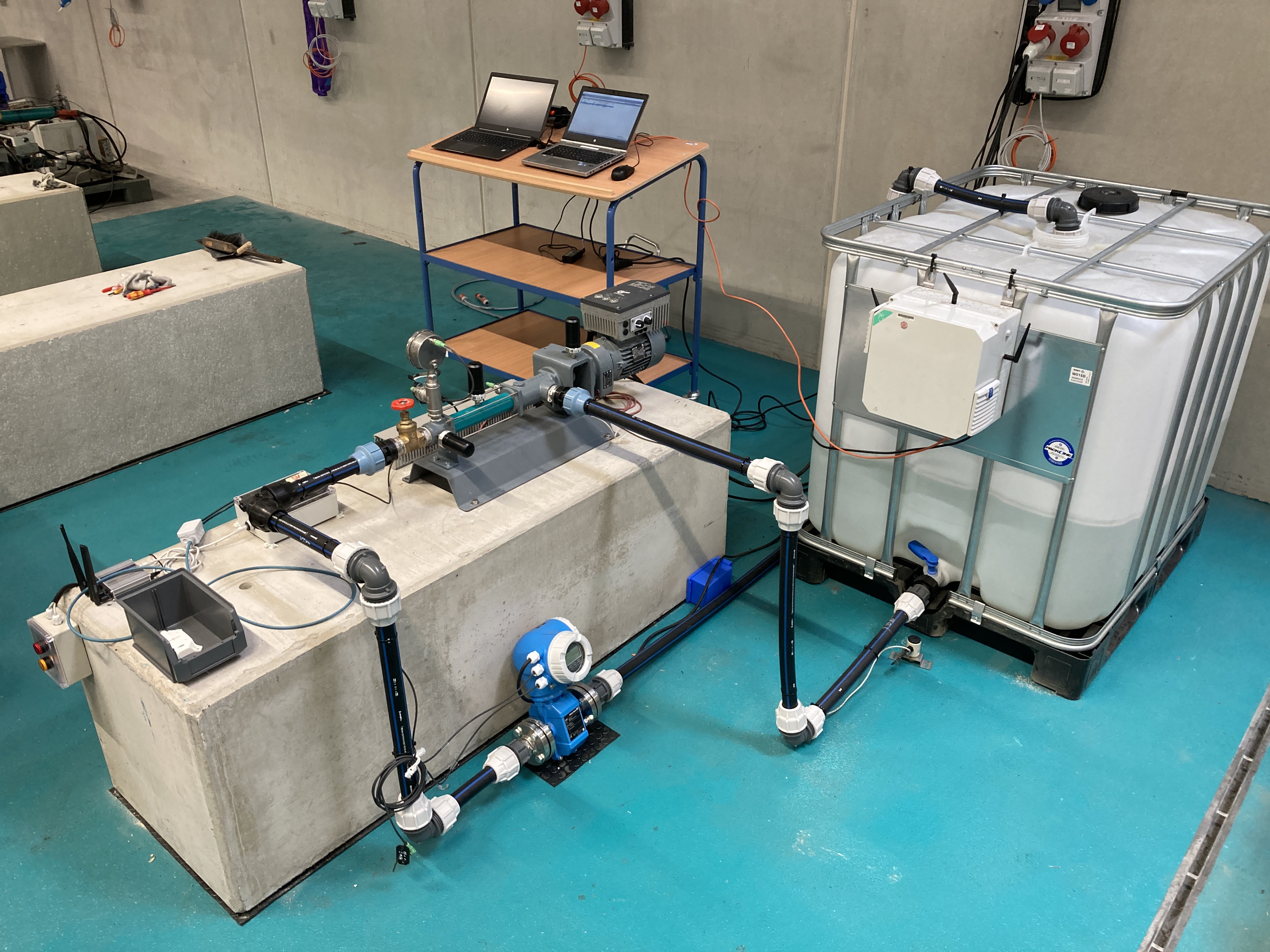}
  \caption{Pump data recording setup}
  \label{fig:netzsch_aufbau}
\end{figure}

The dataset was mainly recorded manually, in part due to the operator's necessary expertise regarding the machine's limit states. For this purpose, the operator sets the pump's specified speed on the frequency converter. The pump's flow rate is throttled with a valve until the specified pressure is reached. Speed and pressure are kept constant for 5 minutes to perform the vibration measurement. The data generation setup can be visualized in Figure~\ref{fig:netzsch_aufbau}.

Initially, label data was recorded at a frequency of 1 Hz. However, challenges in correlating the vibration data with the labels prompted an increase in the sampling frequency to 50 Hz. The vibration data was captured at 6664 Hz for 16384 data points (approximately 2.5 s), with a 13-second transmission gap due to Bluetooth limitations. The final dataset used for this study amounts to approximately 4000 3-axis recordings of 16384 data points each or approximately 170 minutes.

%% file: sections/methods.tex
\label{sec:methods}
This section details the data preprocessing pipeline, the training procedure, the \ac{SNN} architecture, the hyperparameter optimization process, the compression techniques employed, and the evaluation approach.

\subsection{Data Preprocessing}
\label{subsec:preprocessing}

\begin{figure}[t]
    \centering
    \includegraphics[width=\linewidth]{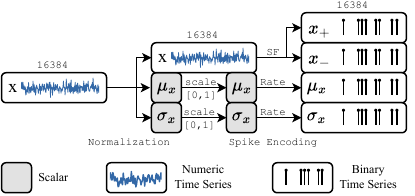}
    \caption{
    Preprocessing applied to each axis (\textit{x, y, z}). 
    After global standardization, a numeric time series undergoes local normalization, yielding a normalized time series and the window's local mean ($\mu_x$) and standard deviation ($\sigma_x$). 
    The normalized time series is encoded into positive ($x_+$) and negative ($x_-$) spikes via \ac{SF} encoding. 
    The local values ($\mu_x, \sigma_x$) are standardized to [0,1] using global statistics and rate-encoded using a Poisson process.
    The pipeline outputs four binary time series per axis. Legend indicates data types.
}
\label{fig:preprocessing_pipeline}
\end{figure}

To prepare the raw sensor data for the \ac{SNN}, a multi-stage preprocessing pipeline was implemented, visualized in Figure~\ref{fig:preprocessing_pipeline}. This pipeline converts the raw 3-axis (\textit{x, y, z}) accelerometer readings, provided as sequences of \texttt{uint16} values within a defined time window ($16384$ samples), into a multi-channel spike train format suitable for neuromorphic processing. The pipeline consists of the following steps, executed sequentially for each time window:

\subsubsection{Standardization}
\label{subsubsec:std}
Initially, the raw data is normalized using the global mean and standard deviation (of each axis), calculated offline from the \textit{entire} training dataset. Next, the mean and standard deviation are calculated again, but this time \textit{locally} for each axis within the \textit{current} time window in order to capture the specific characteristics of the current segment, which would otherwise be lost after normalization. These local statistics are later converted to spikes and added as extra features to the input signal. The sample is then further normalized using the computed \textit{local} mean and standard deviation. This procedure is necessary to ensure the data has the appropriate range to be processed by the subsequent spike encoding step.

\subsubsection{Spike Encoding}
\label{subsubsec:spike_encoding}
The second and final step involves converting the processed numerical data into binary spike trains using a hybrid encoding approach, implemented in the Spike Encoding Framework proposed in \cite{vasilache2025pytorch}. To capture the dynamic fluctuations within the normalized signal, it is encoded into spikes using the \ac{SF} algorithm \cite{SF_SOURCE_kasabov2016evolving}. This temporal encoding scheme generates spikes when the signal exceeds an adaptive threshold, producing separate "+" and "-" spike trains for each axis. The thresholds used in this step are optimized offline, based on the training data distribution, using the spike converter optimization tool in \cite{vasilache2025pytorch}, which finds the optimal threshold that minimizes the reconstruction error. The scalar local statistics (mean and standard deviation) are also encoded using Poisson rate encoding. First, these scalar values are normalized to a range of [0, 1] using parameters derived from the distribution of these statistics across the training set. The resulting normalized value is then rate-encoded over the duration of the time window ($16384$ steps), generating two additional spike trains per axis.

The output of this preprocessing pipeline is a multi-channel spike train. For each of the three input axes (\textit{x, y, z}), four spike channels are generated: \ac{SF} positive-spikes, \ac{SF} negative-spikes, Poisson-coded local mean, and Poisson-coded local standard deviation. This results in $3 \times 4 = 12$ input channels for the subsequent \ac{SNN}.

\subsection{Training Procedure}
\label{subsec:training}
The \ac{SNN} was trained with PyTorch~\cite{pytorch} using a supervised approach targeting both regression and classification tasks simultaneously.

\subsubsection{Data Preparation and Augmentation}
The dataset was split into training ($70\%$), validation ($15\%$), and testing ($15\%$) sets. The long sequences ($16384$ timesteps) were segmented using a sliding window to augment the training data and reduce the sequence length. Segments of length $1024$ timesteps were extracted with a step size of $1$. This increased the number of training samples to approximately $40.6$ million sequences. The validation and test sets, comprising $606$ and $607$ sequences, respectively, were kept in their original length of $16384$ steps, as the recurrent network can handle variable input lengths during inference.

\subsubsection{Batching Strategy}
Given the large size of the augmented training set, a custom batching strategy was employed. Before training, the augmented training set was first shuffled. Then, $1000$ mini-batches containing $9000$ sequences (of length $1024$) were sequentially drawn and used for training updates. This ensures that the entire training set is seen by the model approximately every five epochs.

\subsubsection{Target Variable Scaling}
The continuous output variables for regression (flow, pressure, pump speed) exhibit different value ranges. To address this, the RobustScaler from the scikit-learn library \cite{scikit-learn} was used. This scaler standardizes features by removing the median and scaling according to the interquartile range (IQR), making it less sensitive to outliers compared to standard scaling. The scaler parameters were determined using \textit{only} the \textit{training} set and applied to the entire dataset.

\subsubsection{Loss Function}
\label{subsubsec:loss}
A composite loss function was designed to jointly optimize the regression and classification tasks. 
The Mean Absolute Error ($L_1$ Loss) was chosen for the regression task (Equation~\ref{eq:loss_reg}). It was calculated for each of the three target variables (flow, pressure, pump speed) and summed to form the total regression loss: 

\begin{equation}
\label{eq:loss_reg}
\mathcal{L}_{\text{reg}} = \sum_{i=1}^{3} \left| \hat{y}_i - y_i \right|
\end{equation}

\noindent
For the multi-label classification task (normal, overpressure, cavitation), a weighted \ac{CE} Loss was employed (Equation~\ref{eq:loss_class}). Class weights were calculated offline based on the inverse frequency of each class in the training dataset to counteract class imbalance. 
 
\begin{equation}
\label{eq:loss_class}
\mathcal{L}_{\text{CE}}^{\text{weighted}} = - \sum_{c=1}^{C} w_c \cdot y_c \cdot \log(\hat{y}_c)
\end{equation}

\noindent
where \( C = 3 \) is the number of classes, \( w_c \) are inverse-frequency weights, \( y_c \in \{0, 1\} \) is the actual class, and \( \hat{y}_c \in [0, 1] \) the predicted probability. The calculated \ac{CE} loss was additionally rescaled, multiplying it by 10, aiming to adjust the relative contribution of the classification loss compared to the regression loss during optimization. The final loss (Equation~\ref{eq:loss_total}) was the simple sum of the total regression loss and the rescaled, weighted classification loss:

\begin{equation}
\label{eq:loss_total}
\mathcal{L}_{\text{total}} = \mathcal{L}_{\text{reg}} + 10 \cdot \mathcal{L}_{\text{CE}}^{\text{weighted}}
\end{equation}

\subsubsection{Optimization and Regularization}
The network was trained using the Adam optimizer \cite{kingma2014adam} with a learning rate of $1.224 \times 10^{-2}$ and weight decay (regularization) of $1.712 \times 10^{-6}$. A \texttt{ReduceLROnPlateau} scheduler was employed to adjust the learning rate during training. This scheduler monitored the total loss on the validation set and reduced the learning rate by a factor of $0.3$ if no improvement was observed for $3$ consecutive epochs. Furthermore, Early Stopping was implemented, with a patience of $10$ epochs.

\subsection{Network Architecture}
\label{subsec:architecture}

The model employs a recurrent \ac{SNN} designed to process the 12-channel spike train input described in Section~\ref{subsec:preprocessing} for simultaneous regression and classification.

\subsubsection{Neuron Model}
The neuron model employed is the \ac{LIF}, implemented using the snnTorch library \cite{snntorch}. The dynamics of each \ac{LIF} neuron are governed by its membrane potential, which integrates incoming currents (weighted spikes) and decays exponentially over time towards a resting potential (zero in this implementation). When the membrane potential exceeds a predefined threshold, the neuron emits a spike, resetting its potential.

Key parameters defining the neuron dynamics were determined through hyperparameter optimization (see Section~\ref{subsec:hyperopt}) and kept fixed during training. A Membrane Potential Decay $\beta \approx 0.9$ was used, corresponding to the fraction of the membrane potential remaining after one timestep in the absence of input. A threshold value of approximately $0.96$ was employed.

Training was enabled by using a surrogate gradient function, specifically the fast sigmoid function with a slope parameter of $5$, to approximate the derivative of the spiking mechanism during \ac{BPTT}.

\subsubsection{Recurrent Architecture}
\begin{figure}[th]
    \centering
    \includegraphics[width=\linewidth]{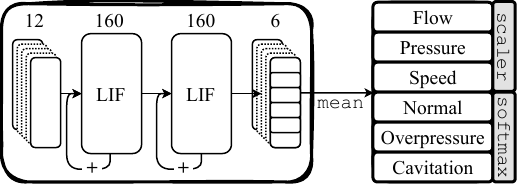}
    \caption{
    Overview of the recurrent \ac{SNN} architecture. 
    It processes 12 input spike channels through two recurrent hidden layers, each with 160 \ac{LIF} neurons. 
    Feed-forward and recurrent connections (indicated by the '+' summation) are present. 
    The final linear layer produces 6 outputs, which are averaged over the sequence timesteps. 
    Regression outputs (flow, pressure, pump speed) are inverse-scaled using the pre-fitted scaler, and classification outputs (normal, overpressure, cavitation) are generated via a Softmax function.
    }
\label{fig:snn_architecture} 
\end{figure}

The network follows a recurrent structure without any bias terms, consisting of an input projection of size $12$, two hidden recurrent \ac{LIF} layers of size $160$, and an output layer of size $6$ (see Figure \ref{fig:snn_architecture}). 
The weights of all linear layers were initialized using a normal distribution with a mean of approximately $-0.048$ and a standard deviation of approximately $0.238$.
The network processes the input sequentially. After the entire sequence ($1024$ timesteps for training, $16384$ for validation/testing) has been processed, the outputs from the final linear layer at each timestep are aggregated by taking their mean. A Softmax activation is applied to the $3$ classification outputs.
A Dropout rate of $0.07$ was applied before the final spiking layer during training to mitigate overfitting.

\subsection{Hyperparameter Optimization}
\label{subsec:hyperopt}
To identify the best hyperparameters for the \ac{SNN} model and training process, we employed the Optuna framework \cite{optuna} to minimize the total loss on the validation set evaluated after training. The optimization utilized a \texttt{MedianPruner} to terminate unpromising trials early, based on intermediate validation loss values compared to the median of intermediate results of previous trials at the same step.

To minimize training time for the hyperparameter optimization, no sliding window augmentation was applied to the training set, resulting in only $2829$ samples. A total of $1000$ optimization trials were conducted, searching learning rate, \ac{LIF} neuron parameters, surrogate gradient slope, weight initialization parameters, the sequence length used during training, and the type of output scaler. An analysis of importance suggested that weight initialization parameters, surrogate gradient slope, and output scaling choice had the highest impact on performance. The optimization process converged to the hyperparameters in Table~\ref{tab:optimized_hyperparams}.


\begin{table}[t]
\centering
\caption{Optimized Hyperparameters}
\label{tab:optimized_hyperparams}
\begin{tabular}{@{}ll@{}}
\toprule
Hyperparameter          & Value         \\
\midrule
Learning Rate                   & $1.224 \times 10^{-2}$ \\
Membrane Decay Factor ($\beta$) & $0.900$                \\
Surrogate Gradient Slope        & $5$                    \\
Neuron Threshold                & $0.959$                \\
Weight Init. Mean               & $-0.048$               \\
Weight Init. Std Dev            & $0.238$                \\
Training Window Length          & $1024$                 \\
Output Scaler                   & \texttt{robust}        \\
\bottomrule
\end{tabular}
\end{table}

\subsection{Compression Methods}
\label{subsec:compression}
To reduce the memory footprint and computational requirements for deployment on resource-constrained hardware, pruning and quantization were applied to the trained \ac{SNN}.

\subsubsection{Pruning}
After training, magnitude-based weight pruning was employed by iteratively removing the smallest absolute weights in increments of $5\%$, with a validation loss tolerance threshold of $0.1$, resulting in a weight sparsity of approximately $25\%$.

\subsubsection{Quantization}
Following pruning, fixed-point quantization was applied to further reduce the model size and prepare it for hardware deployment. Weights were quantized to a signed 18-bit format, with 3 bits allocated for the integer part, based on the maximum absolute weight observed after training. The weight regularization employed during training (Section \ref{subsec:training}) ensured minimal bits were allocated for the integer part by penalizing high absolute weights. Thresholds and Decay Factors ($\beta$) were also quantized to an unsigned 16-bit format, with all 16 bits for the fractional part.
This quantization scheme was applied to the pruned model parameters before evaluating hardware performance metrics. The resulting model has approximately $80,000$ parameters, a memory footprint of $175$\,KB, and a weight sparsity of $32.82\%$ (increased due to quantization).

\subsection{Energy Evaluation}
\label{sec:energy_methodology}


\begin{table}[t]
\centering
\caption{Assumed Energy per Operation (J).}
\label{tab:energy_params}
\begin{tabular}{@{}l S[table-format=1.2e+2] S[table-format=1.2e+2] c @{}}
\toprule
Device & {Synop Energy} & {Neuron Energy} & Source \\ \midrule
CPU x86 (i7-4960X)   & \num{8.60e-09} & \num{8.60e-09} & \cite{degnan2015assessing} \\
CPU ARM (Cortex-A5)  & \num{9.00e-10} & \num{9.00e-10} & \cite{degnan2015assessing} \\ \midrule
Loihi                & \num{2.71e-11} & \num{8.10e-11} & \cite{davies2018loihi} \\
\bottomrule
\end{tabular}
\end{table}

To estimate the energy usage of our model on various platforms, we implemented a methodology inspired by the Nengo framework \cite{bekolay2014nengo}, estimating energy per inference based on synaptic operations and neuron updates:

\begin{equation}
\label{eq:energy}
E = N_{\text{synops}} \times E_{\text{synop}} + N_{\text{neuron updates}} \times E_{\text{neuron}}
\end{equation}
where $E_{\text{synop}}$ is the energy per synaptic operation and $E_{\text{neuron}}$ is the energy per neuron update, specific to the target hardware, summarized in Table~\ref{tab:energy_params}.
The process involves a forward pass of the trained model on the test data (606 samples with 16384 timesteps each), with values averaged across the number of samples. 
The number of neuron updates ($N_{\text{neuron updates}}$) is the number of neurons multiplied by the number of timesteps per inference, which assumes that every neuron is updated at every step. 
For the synaptic operations, the calculation differs between spiking and non-spiking hardware:
    
\subsubsection{Spiking} Because the computation is event-based, the number of synaptic operations is calculated by summing, over all timesteps $t$, layers $l$, and neurons $n$ (in layer $l$) the product of the neuron's spike state $s_{t,l,n}$ (1 if spiking, 0 otherwise) and its number of non-zero outgoing connections $c_{l,n}$. This assumes the spiking platform can leverage connection sparsity:
    \begin{equation}
    \label{eq:synops}
        N_{\text{synops, spiking}} = \sum_{t,l,n} s_{t,l,n} \times \text{c}_{l,n}
    \end{equation}
    where $t$ is timesteps, $l$ layers, and $n$ neurons in layer $l$.

\subsubsection{Non-spiking} It is assumed that all synaptic connections are computed every timestep. Total synops are the sum of all connections multiplied by timesteps per inference.

This methodology has several limitations, as data transfer costs to/from the device are not included, and precision (bit-width after quantization) is not considered. Furthermore, we underestimate the energy cost for the non-spiking devices (x86, ARM) since we assume that each synop and neuron update is one \ac{MAC}, even though in practice, a neuron update might need more, depending on the implementation. Despite this, it provides a comparative estimate of the energy costs across different hardware platforms. Energy estimates for the evaluated model are presented in Section~\ref{sec:results} and summarized in Table~\ref{tab:energy_summary}.

%% file: sections/results.tex
\label{sec:results}
The performance of the trained and compressed \ac{SNN} model was evaluated on the test set (15\% of original recordings, 607 samples). Results for both regression and classification tasks, along with energy estimations, are presented.

\subsection{Regression Performance}

\begin{figure}[t]
    \centering
    \includegraphics[width=\linewidth]{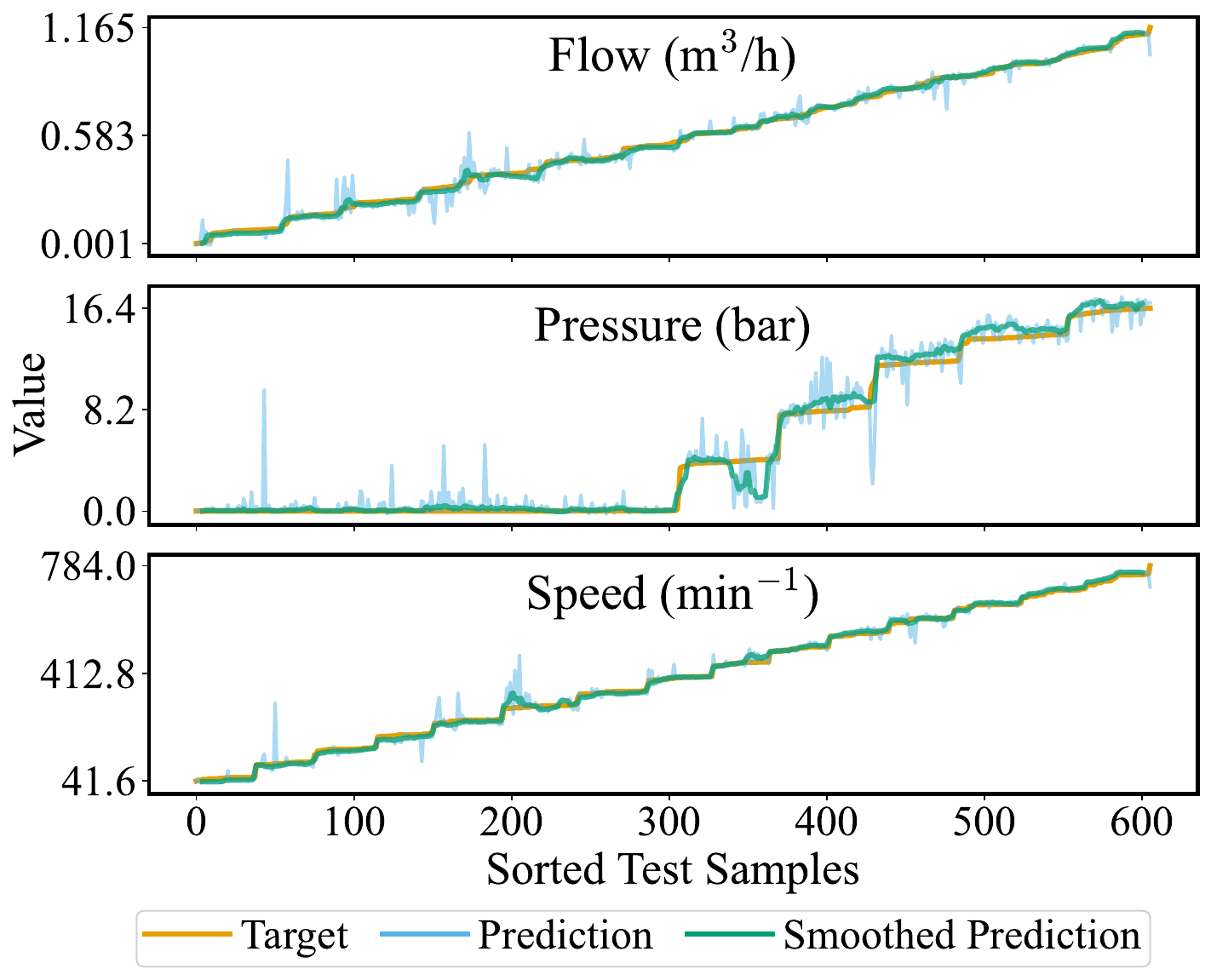}
    \caption{Regression performance on the sorted test set. Targets (orange), raw predictions (light blue), and smoothed predictions (moving median filter with a window size of 10, green) are shown.}
    \label{fig:regression_results}
\end{figure}

Figure~\ref{fig:regression_results} illustrates the model's regression performance for flow ($m^3/h$), pressure (bar), and pump speed ($\text{min}^{-1}$) prediction. The plots show the target values, the raw \ac{SNN} output, and the output smoothed using a moving median filter (window size 10).

To quantify performance, we utilize \ac{MAE}, \ac{MAPE}, and \ac{MRPE}. \ac{MAPE} is defined as:
\begin{equation}
\label{eq:mape}
\text{MAPE} = \frac{100\%}{n} \sum_{i=1}^{n} \left| \frac{y_i - \hat{y}_i}{y_i} \right|
\end{equation}
where $y_i$ is the true value, $\hat{y}_i$ is the predicted value, and $n$ is the number of samples.
\ac{MRPE}, also interpreted as range-normalized error, is defined as:
\begin{equation}
\label{eq:mrpe}
\text{MRPE} = \frac{100\%}{n} \sum_{i=1}^{n} \left| \frac{y_i - \hat{y}_i}{\max(y) - \min(y)} \right|
\end{equation}
where $\max(y)$ and $\min(y)$ represent the maximum and minimum values of the target variable in the test set.

Table~\ref{tab:regression_results} summarizes the quantitative regression results. Smoothing improves all metrics. Comparing smoothed performance to benchmarks: flow achieves a \ac{MRPE} of 0.93\%, within typical industry \ac{FSE} values \cite{sensors_flow}; the pump speed's \ac{MAPE} of 3.12\% exceeds typical industry \ac{MVE} \cite{sensors_rpm1, sensors_rpm2, sensors_rpm3} by 1.12\%; and pressure's \ac{MRPE} of 2.46\% also exceeds its typical \ac{FSE} range \cite{sensors_pressure} by 1.46\%. The high \ac{MAPE} values for pressure are mainly influenced by target values near zero.

\begin{table}[t]
\centering
\caption{Model Regression Performance on Test Set.}
\label{tab:regression_results} 
\begin{tabular}{@{}lccc@{}}
\toprule
Metric & Flow ($m^3/h$)  & Pressure (bar) & Speed ($\text{min}^{-1}$) \\
\midrule
 \acs{MAE} Raw & 0.021 & 0.63 & 10.50 \\
 \acs{MAE} Smooth & 0.011 & 0.40 & 6.90 \\
\cmidrule(l){2-4}
 \acs{MAPE} Raw (\%) & 15.37 & 9237.21 & 4.69  \\
 \acs{MAPE} Smooth (\%) & 4.97 & 2447.47 & 3.12 \\
 Industry \ac{MVE} (\%) & -- & -- & 0.1 - 2.0 \cite{sensors_rpm1,sensors_rpm2,sensors_rpm3}\\ 
\cmidrule(l){2-4}
 \acs{MRPE} Raw (\%) & 1.78 & 3.87 & 1.41 \\
 \acs{MRPE} Smooth (\%) & 0.93 & 2.46 & 0.93 \\
 Industry \ac{FSE} (\%) & 0.1 - 5.0 \cite{sensors_flow} & 0.15 - 1.0 \cite{sensors_pressure} & -- \\ 
\bottomrule
\end{tabular}
\end{table}

\subsection{Classification Performance}

\begin{figure}[t]
    \centering
    \includegraphics[width=\linewidth]{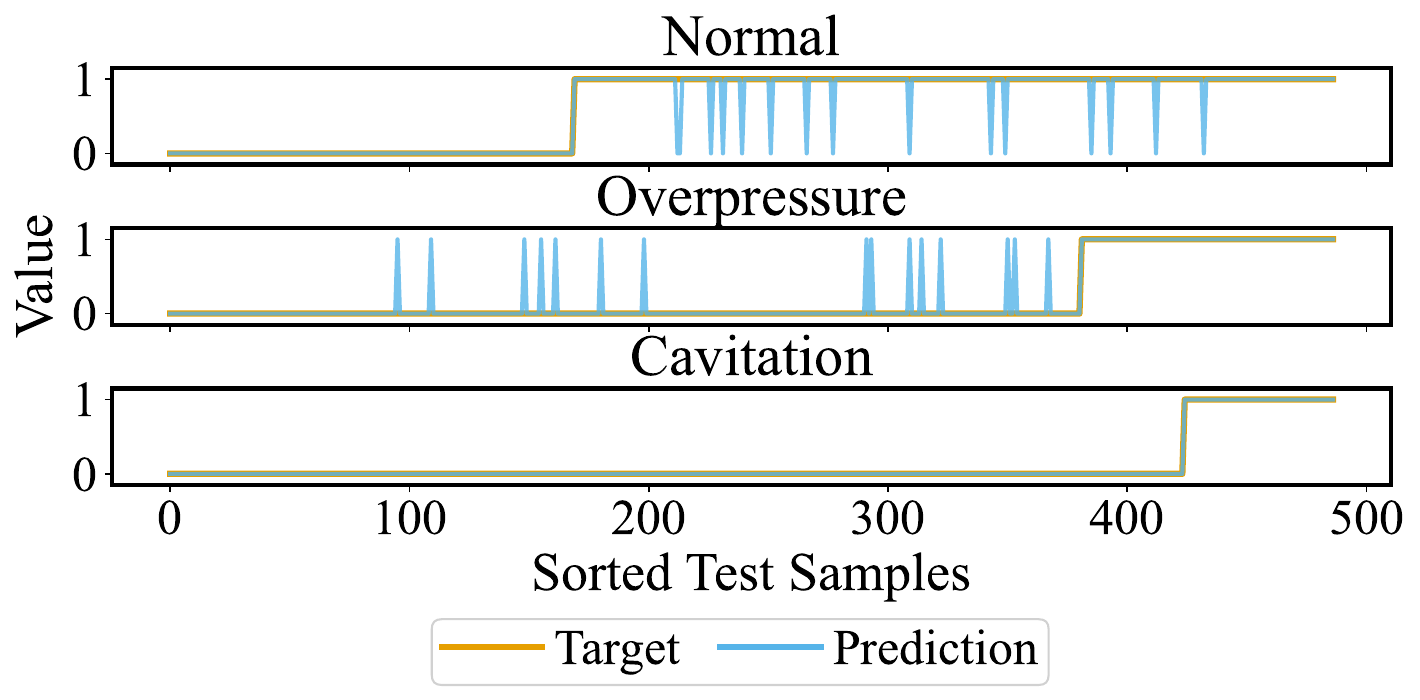}
    \caption{Classification performance on the sorted test set. Targets (orange) and predictions (light blue) are shown.}
    \label{fig:classification_results}
\end{figure}

Figure~\ref{fig:classification_results} shows the classification performance for normal, overpressure, and cavitation conditions. Table~\ref{tab:classification_results} details the classification metrics. The overall accuracy is high (\textgreater97\%). Critically, the \ac{FNR} for overpressure and cavitation is 0.0\%, as the model correctly identified all of these faults in the test set. The \ac{FPR} for overpressure is 3.9\%, meaning some normal samples were incorrectly flagged as overpressure.

\begin{table}[t]
\centering
\caption{Model Classification Performance on Test Set.}
\label{tab:classification_results} 
\begin{tabular}{@{}lcccc@{}}
\toprule
Metric & Normal & Overpressure & Cavitation & Overall \\
\midrule
 Accuracy & 0.9692 & 0.9692 & 1.0000 & 0.9795 \\ 
\cmidrule(l){2-5}
 F1 Score & 0.9758 & 0.9339 & 1.0000 & 0.9699 \\
 \acs{AUC} & 0.9891 & 0.9904 & 1.0000 & 0.9932 \\
\cmidrule(l){2-5}
 \acs{FPR} & 0.0000 & 0.0394 & 0.0000 & 0.0131 \\
 \acs{FNR} & 0.0472 & 0.0000 & 0.0000 & 0.0157 \\
\bottomrule
\end{tabular}
\end{table}

\subsection{Energy Evaluation}

Based on the methodology described in Section~\ref{sec:energy_methodology}, energy consumption estimates are presented in Table~\ref{tab:energy_summary}. The results compare estimates for conventional platforms (x86, ARM) and a neuromorphic platform (Loihi).

Loihi exhibits the lowest estimated energy consumption per inference (\num{3.16e-3} J/inf), significantly lower than the x86 CPU (\num{1.13e+1} J/inf) and the ARM CPU (\num{1.18e+0} J/inf) estimates. It is worth noting that for Loihi most of the energy consumption (86.3\%) is due to the synaptic operations energy, which is dependent on the number of spikes generated within the network (see Equation~\ref{eq:synops}). During a 16384 timestep inference, 530 thousand spikes are generated on average, leading to around 32 spikes within the network in each step. Each neuron thus spikes with an average frequency of around 0.1. It is evident that by reducing this number, or the number of non-zero-weights, the energy consumption would also be reduced, which could be investigated in future work.

\begin{table}[htbp]
\centering
\sisetup{round-mode=figures, round-precision=3, tight-spacing=true}
\caption{Estimated Energy per Inference (J/inf).}
\label{tab:energy_summary}
\begin{tabular}{@{}l S[table-format=1.2e+1] S[table-format=1.2e+1] S[table-format=1.2e-1] @{}}
\toprule
Device       & {Total Energy} & {Synop Energy} & {Neuron Energy} \\
\midrule
CPU          & \num{1.127304e+01} & \num{1.122710e+01} & \num{4.593418e-02} \\
ARM          & \num{1.179736e+00} & \num{1.174929e+00} & \num{4.807066e-03} \\
\midrule
Loihi        & \textbf{\num{3.158786e-03}} & \textbf{\num{2.726151e-03}} & \textbf{\num{4.326359e-04}} \\
\bottomrule
\end{tabular}
\end{table}

%% file: sections/conclusion.tex
\label{sec:conclusion}

This paper demonstrated the application of a recurrent \ac{SNN} for low-power, vibration-based \acf{PM} of a \acf{PCP}. The model successfully performed simultaneous regression of key operating parameters (flow, pressure, pump speed) and classification of pump conditions (normal, overpressure, cavitation).

Key findings include high classification accuracy, notably achieving a zero \acf{FNR} for overpressure and cavitation faults. Regression performance achieved or approached typical industrial sensor standards, supporting the possibility of replacing multiple costly industrial sensors (flow, pressure, pump speed) with a single device that integrates vibration monitoring and local processing with \acp{SNN}. 

Energy evaluation estimates indicated up to 4 orders of magnitude less energy per inference for  neuromorphic hardware compared to conventional x86/ARM platforms, highlighting the potential for considerable energy savings in battery-powered edge devices. It further supports the feasibility of migrating complex data analysis tasks from the cloud directly to the sensor, thus reducing communication overhead and extending operational lifetimes.

Future work should validate these findings through neuromorphic hardware implementation to confirm real-world energy consumption and performance. More aggressive pruning strategies or the integration of spiking frequency minimization within the objective loss could lead to further energy improvements on neuromorphic hardware. Further refinement of the model architecture and frequency-domain preprocessing could improve regression accuracy. Long-term deployment studies are also needed to assess the robustness and reliability of the \ac{SNN}-based \ac{PM} solution in operational environments.